\documentclass[runningheads]{llncs}
\usepackage{graphicx}
\usepackage{marvosym}
\usepackage[marginal]{footmisc}
\usepackage{subfigure}
\usepackage{booktabs}
\usepackage{makecell}
\usepackage{multirow}

\usepackage{amsmath}
\usepackage{float}
\usepackage{tabularray}
\usepackage{diagbox}
\usepackage{amsfonts}
\usepackage{algorithm}
\usepackage{algpseudocode}
\usepackage{hyperref}
\usepackage{cite}
\graphicspath{{figures/}}
\begin{document}
\title{Spatially Resolved Gene Expression Prediction from Histology via Multi-view Graph Contrastive Learning with HSIC-bottleneck Regularization}
\author{Changxi Chi \and Hang Shi \and Qi Zhu \and Daoqiang Zhang  \and Wei Shao}
\institute{School of Computer Science and Technology \\ Nanjing University of Aeronautics and Astronautics }
\authorrunning{}

\maketitle              
\begin{abstract}
The rapid development of spatial transcriptomics(ST) enables the measurement of gene expression at spatial resolution, making it possible to simultaneously profile the gene expression, spatial locations of spots, and the matched histopathological images. However, the cost for collecting ST data is much higher than acquiring histopathological images, and thus several studies attempt to predict the gene expression on ST by leveraging their corresponding histopathological images. Most of the existing image-based gene prediction models treat the prediction task on each spot of ST data independently, which ignores the spatial dependency among spots. In addition, while the histology images share phenotypic characteristics with the ST data, it is still challenge to extract such common information to help align paired image and expression representations. To address the above issues, we propose a Multi-view Graph Contrastive Learning framework with HSIC-bottleneck Regularization(ST-GCHB) aiming at learning shared representation to help impute the gene expression of the queried imaging spots by considering their spatial dependency. Specifically, ST-GCHB firstly adopts the intra-modal graph contrastive learning (GCL) to learn meaningful imaging and genomic features of spots by considering their spatial characteristics. Then, to reduce the redundancy for the extracted features of different modalities, we also add a HSIC-bottleneck regularization term in the GCL to enhance the efficiency of our model. Finally, an cross-modal contrastive learning strategy is applied to align the multi-modal data for imputing the spatially resolved gene expression data from the histopathological images.We conduct experiments on the dorsolateral prefrontal cortex (DLPFC) dataset and observe a significant improvement compared to the existing approaches. These results show the viability and effectiveness of our ST-GCHB for predicting molecular signatures of tissues from the histopathological images.





\end{abstract}
\section{Introduction}
The development of spatial transcriptomics (STs) technology has transformed the genetic research from a single-cell data level to a two-dimensional spatial coordinate system \cite{rao2021exploring}, making it possible to quantify mRNA expression of large numbers of genes within the spatial context of tissues and cells\cite{fang2023computational}. Based on the collected STs data with additional spatial information, several computational studies are developed to explore the spatial expression patterns\cite{sun2020statistical}, identify spatial domain\cite{dong2022deciphering} and infer cell-cell communication\cite{biancalani2021deep}. However, due to the expensive costs of collecting the ST data, the ST technologies haven’t been utilized in large-scale studies while the histopathological images are cheap and easy to acquire, and thus it has becomes an alternative and research hotspot to link the connection between gene expression pattern and the histopathological images. \\
\indent With the development of deep learning and hardware capabilities, many methods have been proposed to predict gene expression through histopathological images\cite{r2}\cite{r21}\cite{r22}\cite{r3}. For example, Bryan et al have proposed the ST-Net to predict the target gene expression of each spot in the whole-slide images \cite{r2}. The HE2RNA model \cite{r21} is a multilayer perceptron (MLP) for predicting the gene expressions from the histopathological images, which allows to perform multitask learning by taking into account the correlations between multiple gene expressions. Other studies include \cite{r22} have proposed an simple but efficient deep learning architecture to predict gene expression from breast cancer histopathology images. \\
\indent Although the above studies show their potential for extracting molecular features from histology slides, they treat the prediction task on each spot of ST data independently, ignoring the spatial dependency among different spots. As a matter of fact, we could identify the spatially continuous patterns of gene expression among different spots in STs data \cite{lopez2022destvi}. To utilize such spatial information to help predict the gene expression values from the histoplogy, Hist2ST\cite{r24} combines the Transformer and graph neural network modules to capture the spatial relations with the whole image and neighbored spots. Jia et al \cite{r23} develop a hybrid neural network that utilizes dynamic convolutional and capsule networks to explore the relationship between high-resolution pathology image phenotypes and gene expression data. All these studies suggest that the spatial information are beneficial in improving the performance for gene prediction tasks. However, it is still challenge to extract common information from the spatial transcriptomics and histology image data. This is because the   histopathological images reveal heterogeneous patterns while the gene expression data usually come with high dimensionality, it is likely that the extracted image features will show poor agreement with their genomic profiles and thus it is important to learn the representations of spots that can both capture the spatial information of the spots and take the association among multi-modal data into consideration. \\
\indent To address the above issues, we propose a multi-view Graph Contrastive Learning framework with HSIC-bottleneck Regularization(ST-GCHB) aiming at learning the shared representation to help impute the gene expression of the queried imaging spots. Specifically, we firstly apply the intra-modal graph contrastive learning method to learn node representation of each modality by considering the spatial information of different spots. Then, a cross-modal contrastive learning module is developed to align the features of the two modalities in the feature space. To reduce the redundancy for the extracted features, a HSIC-bottleneck regularization term is also incorporated in our  ST-GCHB model. We evaluate our method on the dorsolateral prefrontal cortex (DLPFC) dataset, and the experimental results indicate that our method is superior to the existing methods.    

\section{Method}
\subsection{Dataset}
We test our method by performing it on human dorsolateral prefrontal cortex dataset derived from the 10X Visium platform \cite{r2}, we show the number of spots and detected gene in Table 1.  \\
\begin{center}
\begin{tabular}{c|cccccc}
    \hline
    Slice ID  &  151669 & 151670 & 151673 & 151674 & 151675 &151676\\
    \hline
    Num of Spots &3662 &3499 &3641 &3672 &3592 &3460\\
    Total Genes Detected &21235 &21016 &21843 &22437 &21344 &24904\\
    \hline
\end{tabular}
\end{center}
\subsection{Raw Data Preprocessing}
For the gene expression data derived from ST data,  we perform log normalization on all slices and select the top 2000 genes with the highest variance using the tool of Scanpy\cite{scanpy}. When retrieving the images corresponding to the sampling spots, we crop them with the pixel coordinates of the sampling points in the image as the center.

\begin{figure}[t]
    \centering
    \includegraphics[width=12cm]{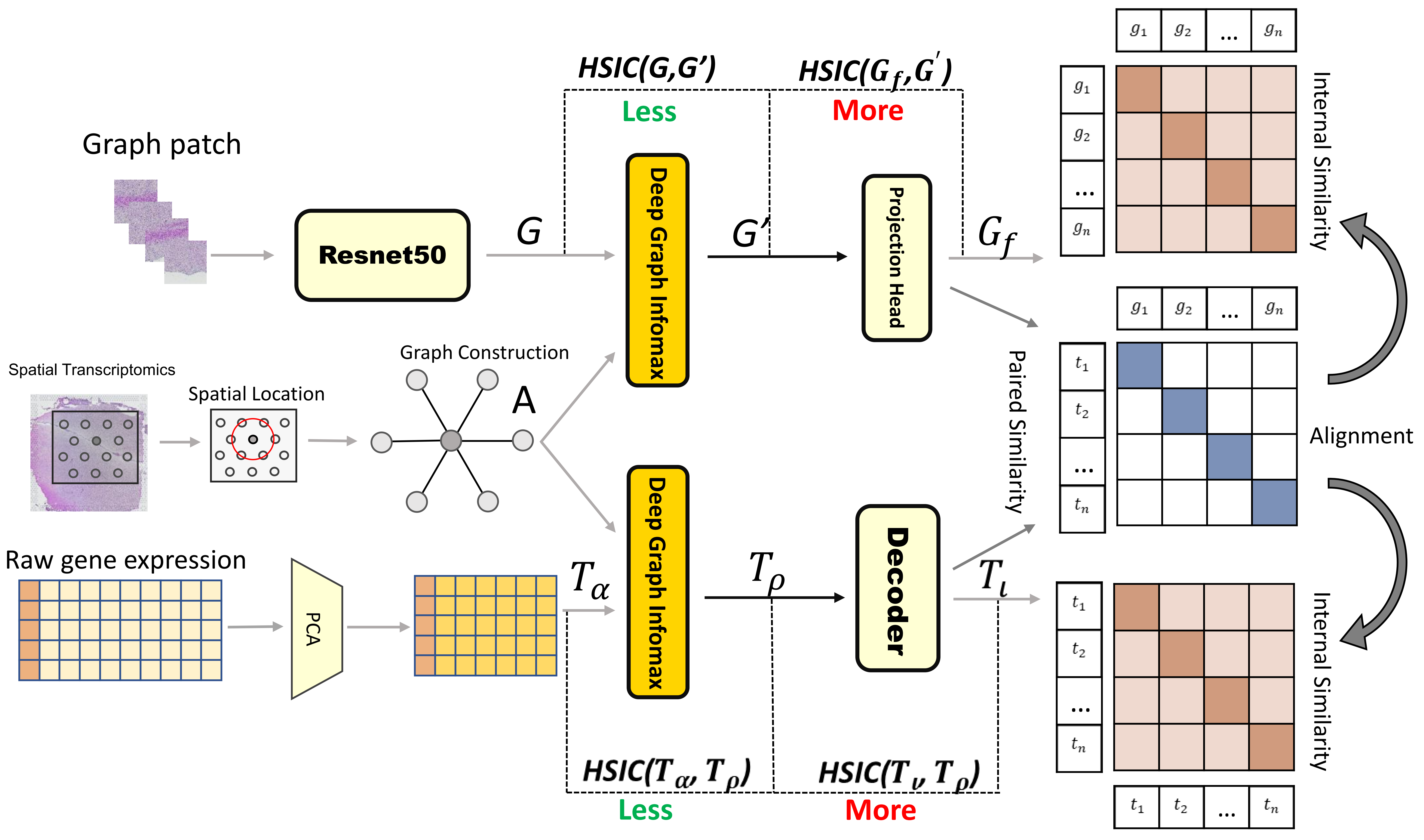}
    \caption{ST-GCHB extracts features from the STs gene expression and adjacency structure of sampling points, incorporating the nHSIC-Bottleneck structure to further eliminate redundant information. Simultaneously, it segments corresponding image patches from H\&E stained images to extract image features. Ultimately, ST-GCHB aligns matching feature pairs in the feature space.}
    \label{fig:enter-label}
\end{figure}
\vspace{-0.5cm}
\subsection{Spatial Adjacency Information}
During the sampling of STs data, our sampling spots are strategically distributed across the STs chip to ensure a uniform spatial distribution. It is widely acknowledged that in biological tissues, the gene expression of a cell is often influenced by its neighboring cells. Consequently, we posit that each sampling spot on the chip is closely connected to its adjacent spots. To reveal the distribution patterns of gene expression more effectively, we introduce the Graph Contrastive Learning framework DGI \cite{r9}.\\
In a singal slice, $A$ represents the adjacency matrix of all sampling spots within that slice. Here, $A_{ij}=1$ if and only if the distance between spot $i$ and spot $j$ is less than $R$ in Euclidean space. $Ta=PCA(RE)\in \mathbf{R}^{n\times k}$, serves as the positive sample, while $Tn \in \mathbf{R}^{n\times d}$, generated by shuffling the rows of Ta, serves as the negative sample. We use the DGI encoder F to derive node representations $Z=F(Ta,A)\in \mathbf{R}^{n\times d}$ and $Z^{'}=F(Tn,A)\in \mathbf{R}^{n\times d}$. The global feature $h$ equals $\sum_{i=1}^{n}Z_{i}$. Our objective is to maximize the similarity between positive sample pairs and minimize that between negative sample pairs, thereby enabling the encoder $F$ to effectively learn node representations capturing graph structures and inter-node relationships, facilitated by the Discriminator $D$. In the image modality, we employ the ResNet50 \cite{resnet50} model with only its last layer being trainable to extract image features, and followed the same training method as described above.
\begin{equation}
    \mathcal{L_{DGI}}=-\sum_{i=1}^{n}[log(D(Z_{i},h))+log(1-D(Z_{i}^{'},h))]
\end{equation}

\subsection{HSIC-Bottleneck}
The IB reveals how networks extract crucial information while filtering out redundant information. In deep learning, IB aim to minimize the MI between the input data X and the intermediate layer data T, while simultaneously maximizing the MI between T and the output data Y. This serves the purpose of feature purification, with the specific objective outlined as follows:
\begin{equation}
    minimize{X,T,Y}(I(X;T)-\beta I(Y;T))
\end{equation}
Unfortunately, accurately estimating probabilities or probability densities accurately is often challenging. One possible approach is to estimate based on sample data, such as Kernel Density Estimation (KDE\cite{r10}), Histogram and Gaussian Mixture Model (GMM\cite{r11}). However, KDE requires great computational resources, histograms face challenges in high-dimensional spaces, and GMM may struggle to capture non-linear relationships.
\\HSIC offers a potential solution to this challenge by providing a parameter-independent measure of probability distribution correlation, with lower computational costs. HSIC assesses the relationship between variables by comparing their mappings in Hilbert space. It is commonly employed to measure the correlation between two random variables, focusing on their interdependence rather than simply evaluating their linear relationship. The unbiased estimate of HSIC is(we assume both \(|A|\) and \(|B|\) equals n):
\begin{equation}
    HSIC(A,B)  =\frac{1}{(n-1)^{2}}Tr(K_{A}JK_{B}J)
\end{equation}
where $K_{A},K_{B}\in \textbf{R}^{n \times n}$ obtained through the Gaussian kernel, and $1$ is an n-dimensional all-ones matrix, $J=I_{n}-1/n$. Based on this, derive the normalized-HSIC(nHSIC), which could effectively prevent the gradient vanishing problem during the training process.
\vspace{-0.3cm}
\begin{equation}
    nHSIC(A,B)  =\frac{HSIC(A,B)}{\sqrt{(1+HSIC(A,A))(1+HSIC(B,B))}}
\end{equation}
Above, we accomplish the objective using nHSIC to delineate the correlation among various representations. Here, $\beta$ is a hyperparameter balancing information compression and preservation. For instance, in our network, a larger $\beta$ makes the network focus more on the correlation between $T_{\alpha}$ and $T_{\rho}$, while a smaller $\beta$ directs the network's attention more towards the information compression between input $T_{\iota}$ and $T_{\rho}$.
\begin{equation}
    minmize_{T_{\alpha},T_{\rho},T_{\iota}}(nHSIC(T_{\alpha};T_{\rho})-\beta nHSIC(T_{\iota};T_{\rho}))
\end{equation}

\subsection{Contrastive Learning}
On STs data, each sampling spot is associated with a histopathological image patch. Considering the extracted embeddings of two modalities as matching pairs, our goal is to align these matching pairs in the feature space while pushing non-matching pairs away from each other. After extraction of the ST-GCHB, we obtain image features matrix $G_{f}\in\mathbf{R}^{B\times d}$ and gene expression features matrix $T_{\iota}\in\mathbf{R}^{B\times d}$, where $B$ represents the batch size and $d$ denotes the feature dimension.\\
Inspired by existing methods\cite{r1,r8,r27}, we enhance the model's performance using contrastive learning. Initially, we calculate the similarity $sim_{cor}=T_{\iota} \times G_{f}^{T}$ between two modalities along with their internal similarities $sim_{spt}=T_{\iota} \times T_{\iota}^{T}$ and $sim_{img}=G_{f} \times G_{f}^{T}$. Based on this, we derive a target matrix. This matrix serves as the alignment direction for the two modalities in the latent space, and we utilize cross-entropy as the final loss function.\\
\begin{equation}
    target=softmax((sim_{img},sim_{spt})/2\cdot \mathcal{T})
\end{equation}
\begin{equation}
    Alignment=mean( CE(sim_{cro},target)+CE(sim_{cro}^{T},target^{T}))
\end{equation}
where $\mathcal{T}$ is a temperature hyperparameter, CE represnets cross entropy loss.
\vspace{-0.5cm}
\subsection{Prediction}
Motivated by the approach proposed by BLEEP\cite{r1} and  SeuratV3\cite{r20}, we employ a trained ST-GCHB model to extract features from the gene expression of the training set, which serves as the database for queries. When the histopathological patches and adjacency matric are inputted for prediction, features are extracted via the model and the k most similar features from dataset are retrieved from the joint space. Finally, we achieve prediction expression through linear combination using indexing.

\begin{algorithm}[H]
\caption{Query steps:}
\begin{algorithmic}[1]
\State Input query H\&E image patch set $Q\in \mathbf{R}^{N\times3\times L\times L}$, where N is the number of patches.
\State nHBEP=train-nHBEP$(A,RE,P,\mathcal{H},\mathcal{G})$//get pretained nHBEP (Encoder $\mathcal{H}$ and image Encoder $\mathcal{G}$)
\State dataset=$\mathcal{H}(A,T_{\alpha})$//make dataset of gene expression embeddings 
\State $Emb_{image}=\mathcal{G}(Q)$//get iamge patch embeddings
\State $Sim$=dataset $\times Emb_{image}^{T}$
\State indices=find\_topk($Sim$)
\State prediction=average($RE$[indices])
\State return prediction
\end{algorithmic}
\end{algorithm}

\vspace{-0.2cm}
\section{Experiments and Results}
\subsection{Correlation of Expression Prediction}
We selected data from 6 tissue slices, with 5 randomly chosen for the training set and the remaining 1 for the test set. Table 1 displays the gene expression prediction results on the DLPFC dataset. Our task is to predict 2000 selected genes and evaluate the correlation between the predicted expression of marker genes (ATP2B4, RASGRF2, LAMP5, B3GALT2\cite{r7}), the top 50 most highly variable genes (HVG), and the top 50 most highly expressed genes (HEG) with the ground truth.\\
ST-GCHB and BLEEP exhibit significant advantages over other methods. This is because these approaches, as generative models, are cursed by dimensionality, making them suitable only for predicting a limited number of gene types (the order of magnitude of $10^2$). They are incapable of predicting a vast number of gene types (the order of magnitude of $10^3$). However, ST-GCHB and BLEEP ingeniously address this issue by aligning and exploring features from different modalities in the latent space. While Hist2RNA and THItoGene consider spatial structure and construct the corresponding adjacency matrix, the high dimensionality of single-cell transcriptomic data may limit the model's ability to handle this spatial information effectively, potentially introducing some complexity or noise and thus compromising the accuracy of the results compared to methods that do not consider spatial structure. \\
The improvement of ST-GCHB over BLEEP is attributed to the utilization of spatial information and the deployment of the HSIC-Bottleneck. Specifically, guided by DGI modules, ST-GCHB leverages spatial positional information of genes within cells or tissues, incorporating spatial correlation into the model to more accurately capture the interactions between genes and histopathological images. Furthermore, the introduction of the HSIC-Bottleneck module further enhances the performance of the model, facilitating feature extraction and representation learning. Consequently, this enables deeper exploration and interpretation of gene expression data.\\
Although some improvement has been achieved in the aforementioned gene prediction tasks, the overall prediction accuracy of all highly variable genes still remains at a relatively low level. It remains challenging to uncover deeper associations between the two modalities.
\begin{table}[t]
    \centering
    \caption{The average correlation among the top 50 
 most highly variable genes (HVG), top 50 most highly expressed genes (HEG), and marker genes (MG) with the ground truth}
    \begin{tabular}{c@{\hspace{1cm}}  c@{\hspace{1cm}}  c@{\hspace{1cm}}  c}
        \hline
        Method &MG & HVG & HEG \\
        \hline
        Hist2RNA  & $0.0536\pm0.0021$ & $0.0208\pm0.0029$ & $0.0754\pm0.0059$\\
        Hist2ST   & $0.0515\pm0.0100$ & $0.1072\pm0.0025$ & $0.0339\pm0.0016$  \\
        THItoGene & $0.0248\pm0.0013$ & $0.1131\pm0.0001$ & $0.0601\pm0.0028$\\
        HisToGene & $0.0692\pm0.0063$ & $0.2893\pm0.0016$ & $0.1026\pm0.0003$ \\
        ST-Net    & $0.0681\pm0.0079$ & $0.2858\pm0.0008$ & $0.1240\pm0.0001$ \\
        \hline
        BLEEP     & $0.1020\pm0.0002$ & $0.3985\pm0.0010$ & $0.3324\pm0.0055$ \\
        \textbf{ST-GCHB} & $\textbf{0.1509}\pm0.0109$ & $\textbf{0.4310}\pm0.0057$ & $\textbf{0.3545}\pm0.0019$ \\
        \hline
    \end{tabular}
\end{table}

\begin{figure}[t]
    \centering
    \includegraphics[width=12cm]{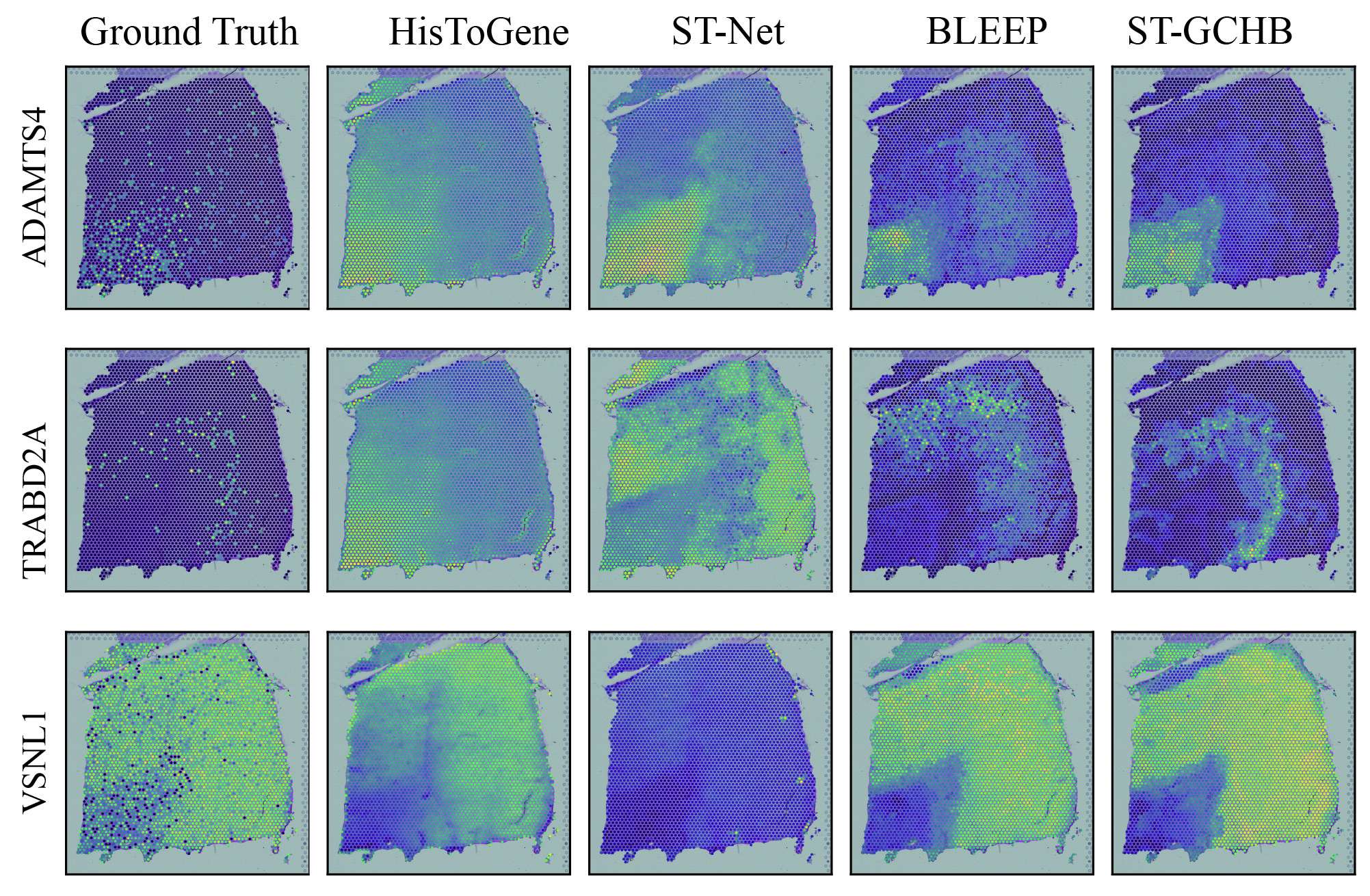}
    \caption{Visualization in spatial coordinates obtained from four methods.}
\end{figure}

\subsection{Ablation Study}
In order to investigate the effectiveness of our method, we conduct ablation study, particularly focusing on exploring the effects of the nHSIC-Bottleneck and graph contrastive learning module. The baseline  represents the complete ST-GCHB. A represents removing the optimization of nHSIC-Bottleneck from the baseline. B represents removing the dimensionality reduction operation of the raw data from the baseline. C  represents the DGI module with the image modality removed from the baseline, and D represents the DGI module with the gene modality removed from the baseline. We evaluate the 4 models on the same dataset above.\\
Model C and Model D consider spatial information from the image and gene modalities respectively, yet they do not outperform BLEEP. However, Model A, as a fusion of Model C and Model D, effectively leverages spatial information from both modalities and achieves promising results. We can infer that unilaterally considering spatial information may cause one modality to accommodate the other, diluting the impact of spatial information. Conversely, symmetrically considering spatial information enables mutual enhancement between the two modalities.\\
Baseline performs better than BLEEP, benefiting from dimensionality reduction. Deep neural networks are often more suitable for learning dense data rather than sparse data. As the raw input, the gene expression matrix itself possesses sparse characteristics. Dimensionality reduction can present its implicit features without the need for subsequent model learning. Baseline, built upon Model A, introduces HSIC-Bottleneck to guide parameter training. Compared to Model A, Baseline achieves significant improvement, demonstrating the the positive significance and effectiveness of HSIC-Bottleneck in feature extraction.
\begin{table}[htbp]
    \centering
    \caption{Ablation Study. Baseline: ST-GCHB. Model A :Baseline without nHSIC-Bottleneck. Model B :Baseline without dimensionality reduction operation. Model C :Baseline without DGI module of image modality. Model D :Baseline without DGI module of gene modality. Cor\textgreater0.3:The number of genes with correlation \> 0.3. AG:average correlation of all predicted expressions.}
    \begin{tabular}{c@{\hspace{1cm}}  c@{\hspace{1cm}}  c@{\hspace{1cm}}  c@{\hspace{1cm}} c@{\hspace{1cm}} c}
        \hline
        Model &MG & HVG & HEG & Cor\textgreater0.3 &AG\\
        \hline
        BLEEP & 0.1177 & 0.3975 &0.3270 &155 &0.0916\\
        Baseline & 0.1617 & 0.4366 &0.3564 &181 &0.1022\\
        \hline
        A & 0.1314 & 0.4188 &0.3470 &160 &0.0960\\
        B & 0.1395 & 0.4031 &0.3211 &163 &0.0953\\
        C & 0.0869 & 0.3814 &0.3012 &144 &0.0887\\
        D & 0.1361 & 0.4099 &0.3328 &169 &0.0991\\
        \hline
    \end{tabular}
\end{table}

\subsection{Experiment Settings}
Our experiments are running with only one Nvidia RTX 3090 Ti GPU (24GB memory) with the AdamW optimizer\cite{r26}. To reduce the training time cost, we pre-cropped the H\&E image patches and conducted lookups based on indices. The experiment used a minibatch size of 16.
\section{Conclusion}
In this paper, we propose a Multi-view Graph Contrastive Learning framework with HSIC-bottleneck Regularization(ST-GCHB) aiming at learning shared representation to help impute the gene expression of the queried imaging
spots. The main advantage of this study is that we consider the spatial information among different spots to help learn the genomic expression from the whole-slide images. The experimental results indicate that our method can not only achieve higher prediction accuracy than the comparing methods but also effectively identify spatial gene expression patterns revealed in STs data. 


\bibliographystyle{unsrt}
\bibliography{reference}

\end{document}